\documentclass[12pt]{article}

\usepackage{scicite}

\usepackage{hyperref}

\usepackage{times}

\usepackage{graphicx}

\usepackage{caption}

\newenvironment{sciabstract}{%
\begin{quote} \bf}
{\end{quote}}

\title{Deep Transfer Learning on Satellite Imagery Improves Air Quality Estimates in Developing Nations}

\author
{Nishant Yadav,$^{1,2}$ Meytar Sorek-Hamer,$^{2,3\ast}$ Michael Von Pohle,$^{2,3}$ \\
Ata Akbari Asanjan,$^{2,3}$ Adwait Sahasrabhojanee,$^{2,3}$ Esra Suel,$^{4}$ \\ Raphael Arku,$^{5}$ Violet Lingenfelter,$^{1,2}$ Michael Brauer,$^{7}$ Majid Ezzati,$^{4}$ \\
Nikunj Oza,$^{3}$ Auroop R. Ganguly$^{1}$\\
\\
\normalsize{$^{1}$Sustainability and Data Sciences Laboratory, Northeastern University, Boston, USA}\\
\normalsize{$^{2}$University Space Research Association (USRA), Mountain View, USA}\\
\normalsize{$^{3}$NASA Ames Research Center, Mountain View, USA}\\
\normalsize{$^{4}$Imperial College London, London, UK}\\
\normalsize{$^{5}$University of Massachusetts, Amherst, USA}\\
\normalsize{$^{7}$University of British Columbia, Vancouver, Canada}\\
\\
\normalsize{$^\ast$To whom correspondence should be addressed; E-mail: msorekhamer@usra.edu}
}

\date{}

\begin{document} 


\baselineskip24pt


\maketitle


\begin{sciabstract}
  Urban air pollution is a public health challenge in low- and middle-income countries (LMICs). However, LMICs lack adequate air quality (AQ) monitoring infrastructure. A persistent challenge has been our inability to estimate AQ accurately in LMIC cities, which hinders emergency preparedness and risk mitigation. Deep learning-based models that map satellite imagery to AQ can be built for high-income countries (HICs) with adequate ground data. Here we demonstrate that a scalable approach that adapts deep transfer learning on satellite imagery for AQ can extract meaningful estimates and insights in LMIC cities based on spatiotemporal patterns learned in HIC cities. The approach is demonstrated for Accra in Ghana, Africa, with AQ patterns learned from two US cities, specifically Los Angeles and New York.   
\end{sciabstract}


\section*{One Sentence Summary}

Estimating air quality in low-resourced developing cities with limited observations is feasible with deep transfer learning.

\section*{Introduction}

Accurately estimating urban air quality (AQ) is vital for effective policy and research. However, not all parts of the world are equally equipped with an adequate ground monitoring network (see figure 1). Cities in low- and middle-income countries (LMICs), such as in sub-Saharan Africa and South-East Asia, severely lack the infrastructure required to monitor AQ \cite{martin2019no}. For instance, of the approximately 15,000 AQ stations reporting to the World Air Quality Index (\href{www.waqi.info}{WAQI}) Project , less than 300 stations are located in Africa. Only 7 out of 54 countries in Africa have real-time AQ monitoring stations \cite{unicef}. In contrast, satellite imagery at increasingly high resolution (HR) is now available globally from governmental and commercial resources. There is, thus, an emerging argument that satellite imagery combined with the latest advances in machine learning (ML) methods - particularly computer vision - may offer a promising alternative approach when traditional methods prove inadequate \cite{zhu2017deep, yuan2020deep, rolf2021generalizable}. Computer vision is a subfield of ML that enables computers to derive meaningful information from digital images and other visual inputs.

Traditional methods for developing city-wide AQ maps can be categorized into physical and empirical approaches. Physical models, such as CMAQ \cite{us_epa_cmaq:_2016}, can provide detailed insights into the physical-chemical processes of the diffusion and transformation of multiple pollutants and present the direct linkage between pollutant emission and air pollution. However, these chemical transport models are coarse-grained (4-12 km spatial resolution) and dependent on a priori knowledge with regard to emissions, which may not be available. Moreover, the underlying physics driving the emissions may not be known understood. On the other hand, empirical models demonstrate the relationships between dependent and multiple independent variables based on historical data. One of the most widely used regression methods in environmental health studies is land-use regression (LUR) modeling \cite{lee2017land, jin2019land}. LUR models solve multiple regression equations that map sample locations and different environmental variables. It requires gathering locally measured data on traffic, weather, land use, and population density, among others. The resulting models can predict AQ levels at unmeasured locations with a high spatial resolution ($\sim$ 100 m). However, data collection for LUR models is time-consuming and may even be prohibitive if requisite data is unavailable. Thus, LUR models are available for only a handful of urban regions and developed on an annual basis. 

Although satellite-based data-products (e.g., Aerosol Optical Depth) have been used to calculate AQ indicators in the past \cite{van2006estimating, just2015using}, in this work, we posit a different approach to inferring air quality from satellite data. We hypothesize that with the availability of sub-meter resolution satellite imagery, next-generation data-driven AQ models might infer pollution information directly from the images using computer vision techniques. The visual features in HR satellite imagery (e.g., parks, roads, industries) indicative of air quality can be distinguished and learned. For example, a satellite image containing a dense road network may indicate a high concentration of air pollutants such as NO2; an image comprising a green cover may suggest the opposite. The latest advances in ML, especially Deep Learning (DL) \cite{guo2016deep, lecun2015deep}, make such a task particularly amenable. Convolutional neural networks (CNN) can learn to identify such features from large training data and have shown promising results in this regard \cite{gu2018recent, pinaya2020convolutional}. Within remote sensing, too, DL methods have been applied to problems such as land use and land cover classification \cite{zhang2019joint}, predicting economic indicators from nightlight imagery \cite{jean2016combining}, and crop estimation \cite{koirala2019deep}. 

\section*{DeepAQ Model - Unsupervised Transfer Learning}

Although DL models, which are data-intensive by design, perform well on regions and problems where large training data is available, generalization to newer, often data-poor areas remains challenging \cite{NIPS2017_10ce03a1}. It limits the direct application of ML/DL techniques to satellite imagery over regions - usually high-income and developed - with abundant ground-level data for model training and validation. We overcome this challenge using a transfer learning approach \cite{zhuang2020comprehensive}. In ML, transfer learning is the idea that knowledge gained while solving one problem can be applied to a different but related problem. Depending on how much (labeled) training data is available over the actual area of interest,  transfer methods can be divided into supervised or unsupervised (see Supplementary Material for more details). 

In this work, we propose an unsupervised transfer learning approach (DeepAQ hereafter) and demonstrate it for the city of Accra in Ghana, Africa. The DeepAQ model performs two steps: first, a CNN-based model is trained to estimate air quality at 200 $m$ resolution (in terms of annual average NO2 levels) over cities with sufficient training data. Los Angeles and New York City (NYC) are chosen as two candidate cities. The two cities are chosen because of 1) availability of labeled data, and, more importantly, 2) a wide distribution of $NO_2$ levels and associated patterns. In the second step, the DeepAQ model is transferred to Accra in a completely unsupervised setting (see Methodology). Our team collected AQ data in Accra for model validation across 130 stations for a year \cite{clark2020high}. As an initial \textit{proof-of-concept}, we also demonstrate the performance of our DeepAQ model by transferring a trained model from LA to NYC, where sufficient labeled data is available for both cities for model validation.


\textit{Satellite Imagery}: Over 300 raw satellite images from MAXAR WorldView2 (WV2) are used in this study to train and validate the DeepAQ model. WV2 was launched in 2009 as part of the MAXAR satellite constellation. It is a sun-synchronous satellite located 770 km from Earth. Images are produced in eight spectral bands in the VIS–NIR range (400 nm–1040 319 nm) with a spatial resolution of 2 $m$ (rescaled to $\sim$ 2.5 $m$ to better match the target data resolution). A single raw image roughly spans an area of 400 $km^2$. In this work, only the visible (RGB) bands are used. From each image, multiple 80 x 80 patches are extracted, such that each patch covers a 200 $m$ x 200 $m$ region on the region to match the target data resolution. In total, approx. 40,000, 16,000, 28,000 image patches are generated for LA, NYC and Accra respectively based on the availability of satellite imagery. 

\textit{Target Data}: The mean annual $NO_2$ target data ($\mu g / m^3$) at 200 $m$ resolution is available from (2010) LUR models for LA and NYC \cite{bechle2015national}. LUR models solve multiple regression equations that map sample locations and various environmental variables. It requires gathering locally measured data on traffic, weather, land use, and population density, among others. LUR data is only available for select few cities as LUR models require substantial time and effort. They commonly serve as inputs for environmental policy and research as it is not possible to obtain station data at such high resolution. 

For Accra, a team of researchers associated with this work collected $NO_2$ ground level data across 135 different locations for a year \cite{clark2020high}. Although in DeepAQ, no labeled data from Accra is used, the 135 datapoints collected serve to validate the predictions over Accra.

\textit{Road Network}: Locations near (major) roads have higher $NO_2$ levels because of the strong correlation with vehicular emission. Based on this understanding, the distance of each target data point from a major road is calculated and fed as an input to the DeepAQ model. The road network information for the three cities is obtained from OpenStreetMap, which is a provider of freely available global geographic database. 

\textit{Data Pre-Processing}: Raw satellite data need to be processed before use. In this work, radiometric correction, followed by filtering for cloud cover was performed on the satellite images. Finally, the images were normalized between the range of $(-1,1)$ before patch extraction.

The problem is formulated as a regression task. The goal is to predict the mean annual $NO_2$ level ($\mu g/m^3$) over an urban region given the corresponding satellite imagery. At each time, the DeepAQ model takes as input two image patches - one from the source city (labeled) and another from the target city (unlabeled). During training, the model learns to generate domain-invariant features for the target city image patches (see section 2.3 for details) while learning to predict $NO_2$ levels for the source city. At the inference, the trained model is used to predict the $NO_2$ levels over the target city using the domain-invariant feature embeddings learned. 

We perform two experiments. The first \textit{proof-of-concept} demonstrates the utility of the DeepAQ model by considering only LA and NYC as the source and target cities, respectively. The goal is to the validate how the model performs by evaluating against the large labeled data available for NYC (not used during training). Once the model efficacy is established, the actual analysis is performing using both LA and NYC as (labeled) source cities and Accra as the (unlabeled) target city. A standard ResNet-34 \cite{he2015deep} based convolutional neural network (CNN) is used as a baseline with no domain adaptation. Two metrics are used to evaluate to performances - standard deviation normalized root mean squared error ($\sigma-rmse$) and coefficient of determination ($R^2$). 

\section*{Results}

The main results are presented in figures 2 and 3. In terms of point metrics, the DeepAQ model substantially outperforms the baseline CNN (Table 1). In the case of NYC, we observe the baseline model fails to capture the spatial distribution of $NO_2$ values, especially over the Manhattan region with high $NO_2$ levels (figure 2). The DeepAQ models not only captures the regions of high $NO_2$ values much better, but is also able to localize distinct regions near roads. However, DeepAQ, too, misses out on areas with extremely high $NO_2$ values. A possible explanation is that such high values ($\geq$ 100 $\mu g/m^3$) are not observed in LA and the model is bounded by the maximum observed value. The DeepAQ model achieves a substantially lower $\sigma-nrmse$ (38\%) and roughly 3 times higher $R^2$ score compared to the baseline (supplementary figure S4). It shows the DeepAQ model is able to successfully transfer features from the source city to the target city. We further analyze this aspect during model interpretation.

\begin{table}
\begin{center}
\begin{tabular}{|l|c|c|c|c|}
\hline
City &
\multicolumn{2}{c|}{NYC} &
\multicolumn{2}{c|}{Accra}\\
\hline
Method & $\sigma-nrmse$ & $R^2$ & $\sigma-nrmse$ & $R^2$\\
\hline\hline
DeepAQ & \textbf{0.525} & \textbf{0.642} & \textbf{0.665} & \textbf{0.492}\\
Baseline & 0.844 & 0.243 & 0.981 & 0.275\\

\hline
\end{tabular}
\end{center}
\captionsetup{labelfont=bf}
\caption{Results. DeepAQ compares favorably to the baseline model which is a simple CNN (resnet-34) without domain adaptation. $\sigma-rmse$ refers to root mean squared error normalized by the standard deviation $\sigma$.}
\end{table}

For Accra, the baseline CNN trained on LA and NYC (without transfer learning) resulted in an $R^2$ score of 0.0. In other words, it fails to capture any distribution over the region. In comparison, DeepAQ results in an $R^2$ score of 0.49 and the predicted distribution overlaps with the actual distribution except for regions of extremely high $NO_2$ values (supplementary figures S5 and S6). 

To summarize, the DeepAQ model succeeds in making reasonably well prediction over unseen regions. However, extremely high and low values are not predicted well. It may be possible to overcome this limitation by using a diverse set of source cities with different AQ distributions. In that case, the model may learn to associate a broader range of values with the urban features identified. 

\subsection*{Model Interpretation}
The goal of the DeepAQ model is to learn high-level features that are indicative of the $NO_2$ levels, for example, roads, freeways, parks. To interpret the model output, we use a pixel attribution method called Grad-CAM \cite{Selvaraju_2017_ICCV}. The method outputs a relevance score for the input image pixels that maximally contributed to the DeepAQ output against the input image. We select two sets of images corresponding to high and low levels of $NO_2$ and generate the pixel-level saliency maps using Grad-CAM for each image (figure 4). We observe that the model is able to identify semantically meaningful features such as freeways in case of high $NO_2$, i.e., the 'freeway' pixels contribute more to the DeepAQ output. Similarly in areas of low $NO_2$, the model is able to identify green areas such as forests and parks. The ability to learn semantically meaningful, and potentially generalizable, features is critical for domain adaptation.

\section*{Conclusion}

Traditional methods for high-resolution AQ mapping are built manually using locally available data. It is difficult to scale them beyond the regions they are built for. Low- and middle-income countries (LMIC) that include some of the most polluted cities lack the resources for direct AQ monitoring. Increasingly fine satellite imagery combined with deep learning methods offer a scalable, automated approach to AQ monitoring making them particularly appealing for LMICs lacking ground data. This work proposes a new approach that entails developing data-driven AQ models using sub-meter satellite imagery that can be transferred to LMIC cities using patterns and insights learned from data-rich cities in the developed parts of the world. Such high-resolution AQ models / maps can facilitate informed decision-making in environmental health policy and research at local scale in the fastest growing urban regions (in LMICs). However, we must not lose sight of the importance of installing high-quality AQ sensors in LMIC cities as the validity of input-output relationship in deep learning models is critically dependent on accurate and reliable estimates of the pollutants of interest.

\bibliography{scifile}

\bibliographystyle{Science}

\section*{Acknowledgments}
This work was supported by the Pathways to Equitable Healthy Cities grant from the Wellcome Trust [209376/Z/17/Z], NASA ARC ARIA Award, USRA IRAD Award, NSF CRISP 2.0 Type 2 INQUIRE Award (1735505) and the NSF INTERN Award. We thank the Pathways to Equitable Healthy Cities team and Wellcome Trust, NASA, USRA, and NSF for the financial and professional support, the NEX group at NASA ARC for supporting our work, Prof. Michael Jerrett (UCLA) for the data and Dr. Jason Su (UC Berkeley) who helped develop the land use regression model in LA, as well as Maxar for providing access to their satellite imagery.

\section*{Author Contributions}
Conceptualization: NY, MSH, MVP, AAA, AS, ARG \\
Methodology: NY, MSH, MVP, AAA \\
Investigation: NY, MSH, MVP, AAA \\
Visualization: NY, VL, ES \\
Funding acquisition: NY, MSH, NO, ARG \\
Supervision: MSH, NO, ARG \\
Data Acquisition: MSH, RA, MB, ME \\
Writing – original draft: NY, MSH, ARG \\
Writing – review and editing: NY, MSH, MVP, AAA, AS, ES, RA, VL, MB, ME, NO, ARG \\

\newpage


\begin{figure}[!ht]
\begin{center}
\includegraphics[width=1.0\linewidth]{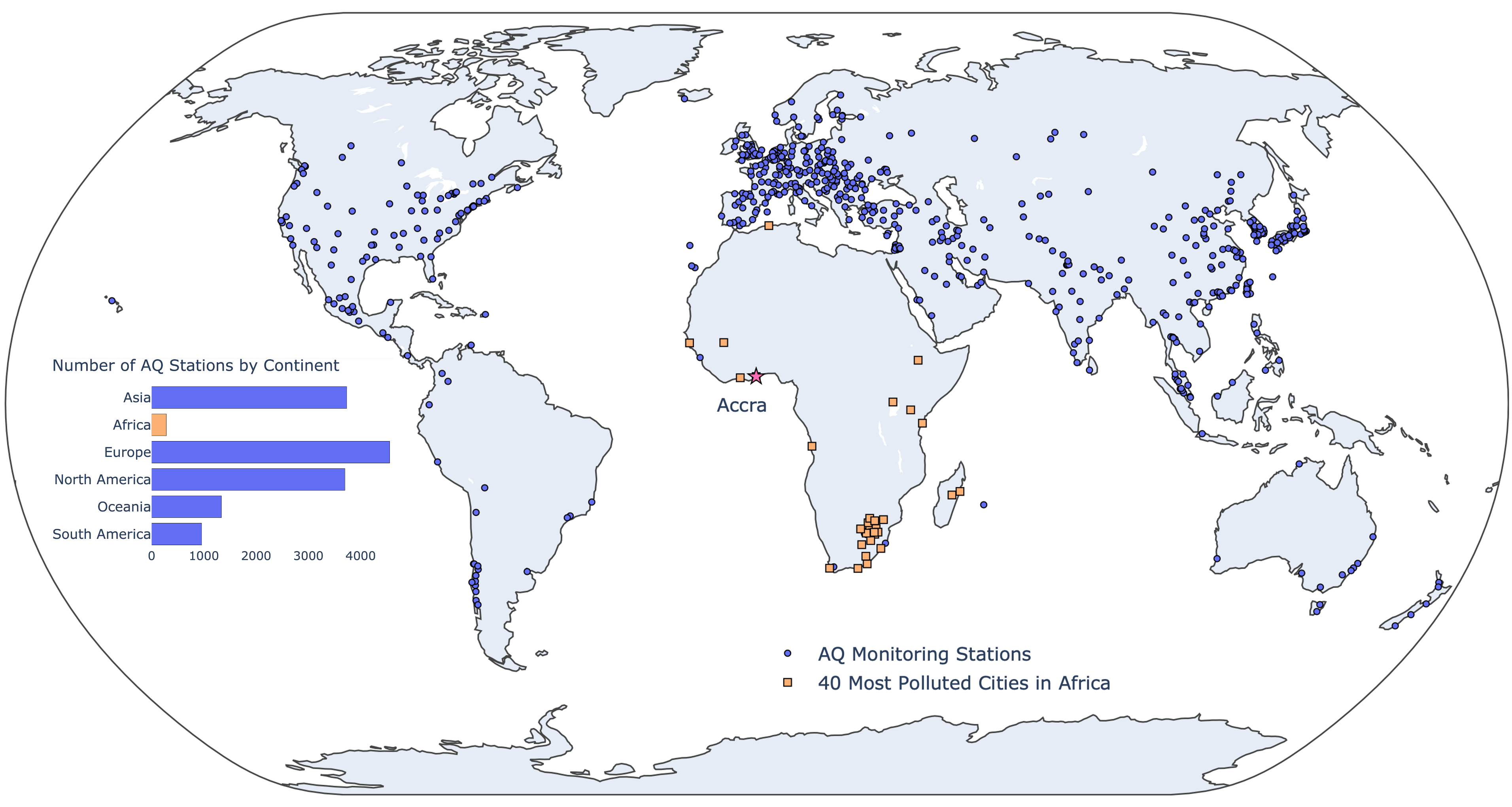}
\end{center}
   \caption{Skewed distribution of air quality (AQ) monitoring stations across the globe. The figure show 681 cities reporting AQ data to the World Air Quality Index (\href{www.waqi.info}{WAQI}) Project through public, private and citizen efforts. Out of approximately 15,000 stations, less than 300 are located in Africa. The third most polluted city in Africa - Accra, Ghana - has only one monitoring station contributing to WAQI compared to 20+ stations for New York City, as an example.}
\end{figure}

\begin{figure}[!ht]
\begin{center}
\includegraphics[width=1.0\linewidth]{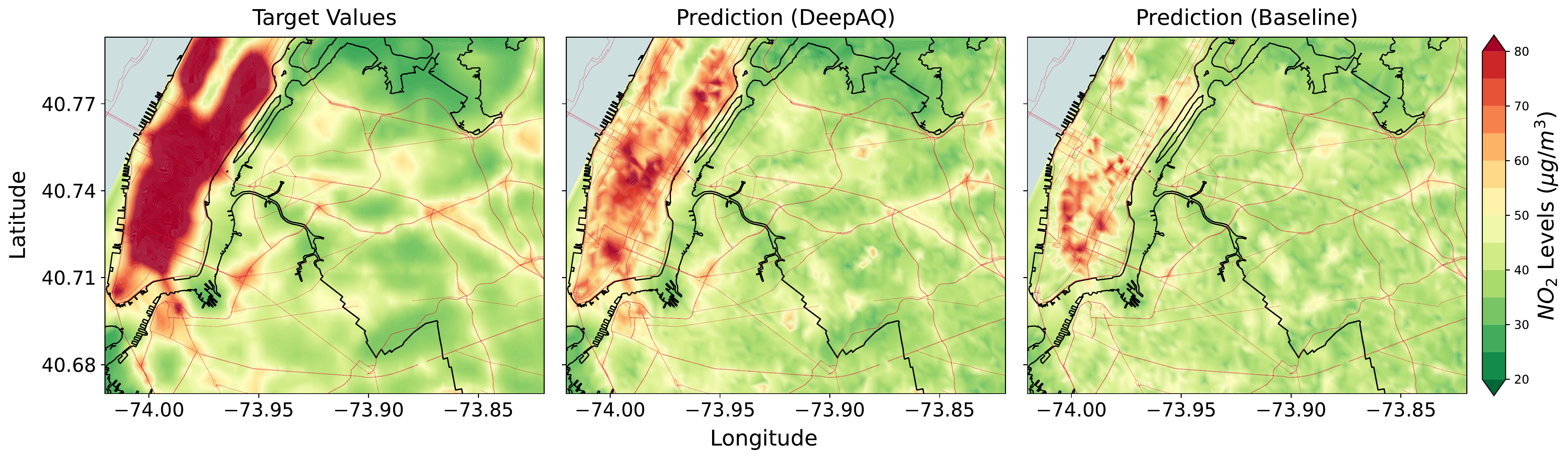}
\end{center}
   \caption{Mean annual $NO_2$ levels ($\mu g/m^3$) over New York City (NYC) as estimated by DeepAQ, and compared with the baseline Resnet-34 model (trained only on Los Angeles (LA) data with no transfer learning). In this case, LA acts as the source city (domain). Compared to the baseline, the DeepAQ model is able to better identify regions of high $NO_2$ levels, such as in the Manhattan area of NYC (upper left). The road network is superimposed on top (red lines) for reference. See supplementary figure S3 for the case when no road network information is provided to the model.}
\end{figure}

\begin{figure}[!ht]
\begin{center}
\includegraphics[width=1.0\linewidth]{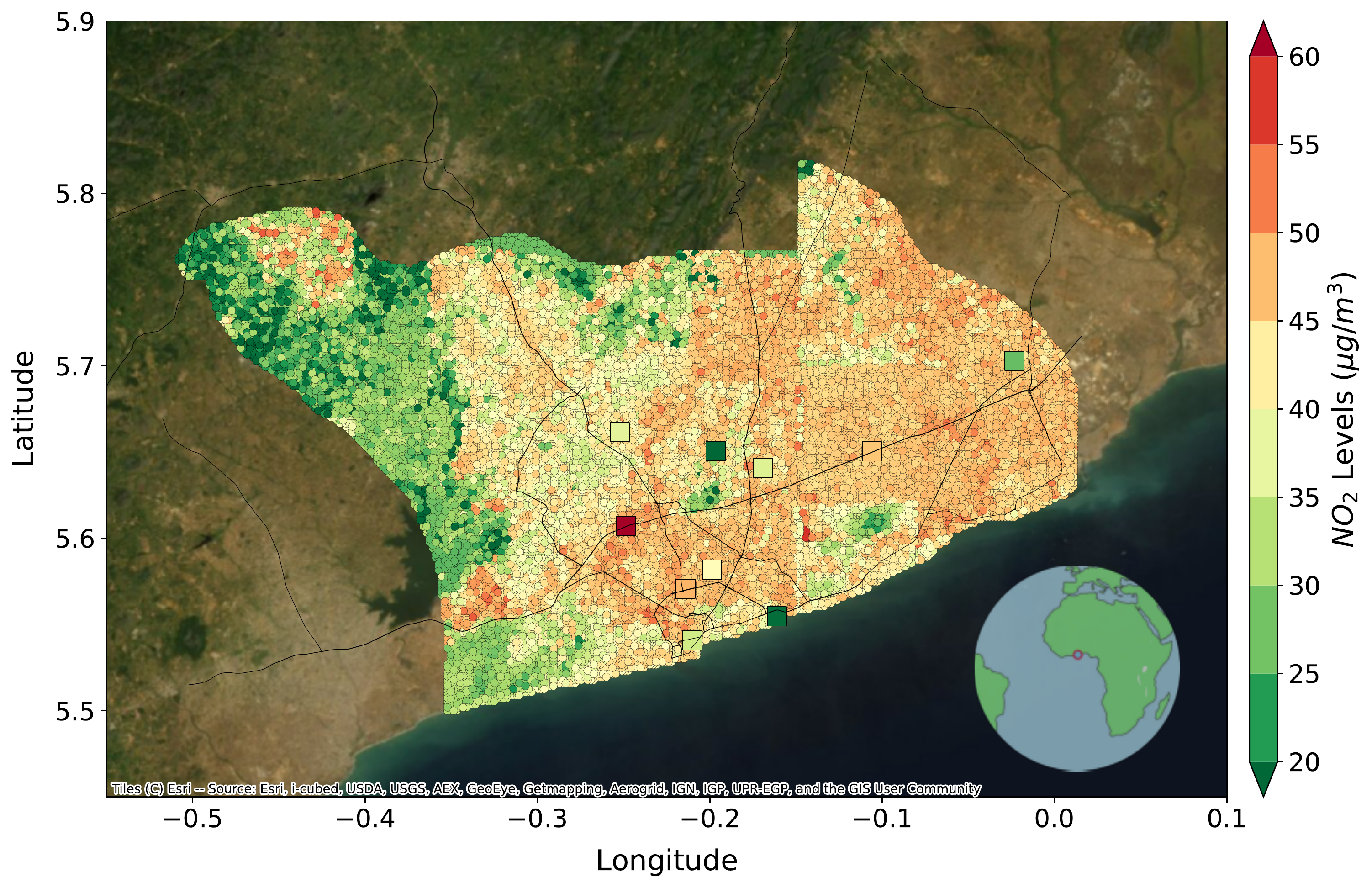}
\end{center}
   \caption{Mean annual $NO_2$ levels ($\mu g/m^3$) over Accra, Ghana, as predicted by DeepAQ.The square dots point to the fixed monitoring stations measuring $NO_2$ levels across Accra. The DeepAQ model is able to capture the spatial distribution of $NO_2$ over Accra with higher levels observed in the city center and near the highways. See supplementary figure S2 for the case when no road network information is provided to the model.}
\end{figure}

\begin{figure}[!ht]
\begin{center}
\includegraphics[width=1.0\linewidth]{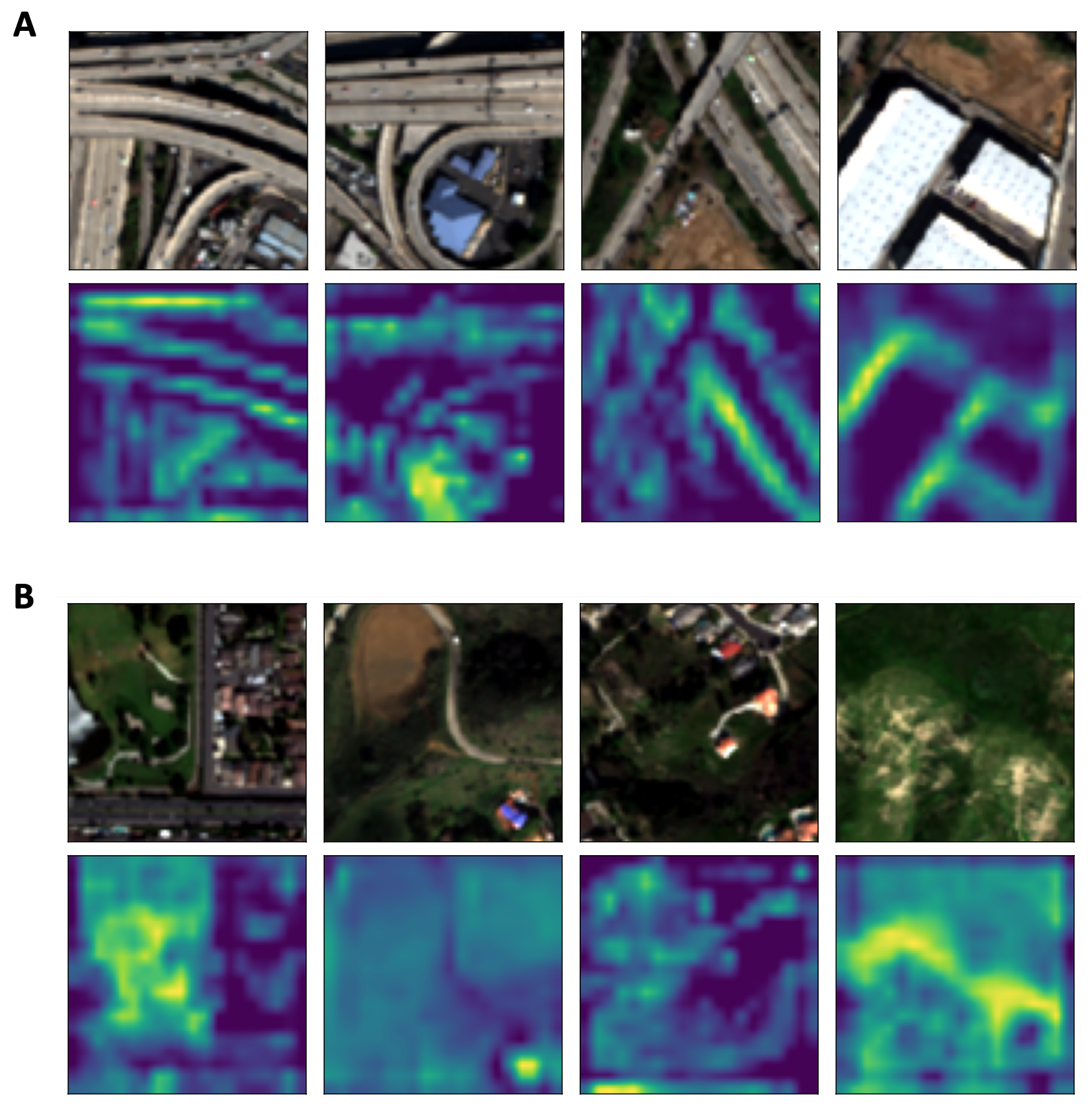}
\end{center}
   \caption{Interpreting the DeepAQ model. Using the Grad-CAM method, regions of the input that are considered important by the model for prediction are visualized. The top row in (A) and (B) contains four satellite image patches from Los Angeles with high and low $NO_2$ levels ($\mu g/m^3$), respectively. The bottom row contains the pixel-importance maps generated using Grad-CAM. Pixels in yellow refer to the pixels that maximally contributed to the DeepAQ output against that image patch. The DeepAQ models is able to learn meaningful high-level features such as freeways and green cover and relate them to different levels of $NO_2$.}
\end{figure}


\clearpage
\section*{Supplementary Materials}

\section*{Methodology}

\subsection*{Transfer Learning}
We define transfer learning using the framework in \cite{pan2009survey}. A domain $\mathcal{D} = \{\mathcal{X}, \mathcal{P(X}\}$ consists of a feature space $\mathcal{X}$ and a marginal probability distribution $\mathcal{P(X)}$. Given a domain, a task $\mathcal{T} = \{\mathcal{Y}, f(\cdot)\}$ consists of a label space $\mathcal{Y}$ and a predictive function $f(\cdot)$ which models $\mathcal{P}(y|x)$ for $y \in \mathcal{Y}$ and
$x \in \mathcal{X}$. Given a source domain $\mathcal{D_S}$ and learning task $\mathcal{T_S}$, and a target domain $\mathcal{D_T}$ and learning task $\mathcal{T_T}$ , transfer learning aims to improve the learning of the target predictive function $f_T (\cdot)$ in $\mathcal{T_T}$ using the knowledge from $\mathcal{D_S}$ and $\mathcal{T_S}$.

\subsection*{Domain Adaptation}

Based on whether $\mathcal{D_S} \neq \mathcal{D_T}, \mathcal{T_S} \neq \mathcal{T_T}$ or both, transfer learning can be categorized into different sub-types. A comprehensive overview of transfer learning approaches is presented in \cite{pan2009survey}. A common use case arises from the phenomenon known as dataset shift or \textit{domain shift}, where the marginal probability distribution of the feature space $\mathcal{X}$ in source domain $\mathcal{P_S(X)}$ is different from the target domain, $\mathcal{P_T(X)}$. As a result, the models trained on the feature representations from one domain do not generalize well on novel (target) datasets (domains). The typical solution is to use a pre-trained model and fine-tune using task-specific datasets. However, it might be prohibitively difficult to obtain enough labeled data to properly fine-tune the large number of parameters employed by DL models. Unsupervised Domain adaptation (UDA) is the problem for overcoming such a domain shift and facilitate improved generalization of DL models to novel 'unlabeled' datasets. The more successful UDA methods attempt to learn feature representations that are domain-invariant. In other words, the DA model maps the source and target features into a common feature space to ensure that the model cannot distinguish between the training and test domain data samples. By doing so, the model is expected to perform (generalize) better on feature embeddings, and ultimate the downstream task, of the target domain. 

\subsection*{Proposed Model}

The proposed UDA model (DeepAQ) is based on an adversarial training approach, first proposed by Ganin et al \cite{ganin2015unsupervised}. The model (see Fig. S1) comprises of three components – a convolutional feature encoder (G), a fully connected classifier or regression decoder (D) and an auxiliary domain classifier (critic, C). In the first step, the encoder takes the labeled source and unlabeled target domain images as inputs and generates feature embeddings. The critic C then takes the feature embeddings generated by G and attempts to classify them as coming either from source or target-domain. The encoder G is then trained with an additional adversarial loss (with gradient reversal) that maximizes C’s mistakes and thus aligns features across domains. Further, during training only the labeled source images are fed through to the decoder D which predict the mean annual $NO_2$ level over that satellite image patch using a Euclidean loss. At inference, the domain-invariant features for the target domain learned by the feature encoder G are passed through decoder D for prediction the mean annual $NO_2$ levels for the target domain image patches. We adapt the original approach in Ganin et al. by adding a strong regularization in the encoder (G), which substantially improved the performance (over 10\% for synthetic datasets such as MNIST). Empirically, we observed that the regularization allows for learning more robust and potentially more generalizable features that are equally discriminative at the same time.

\subsection*{Baseline}
In DeepAQ, a pre-trained ResNet-34 model is used as the feature encoder component (G). For the baseline model, the critic (C) responsible for \textit{transferring} or adapting the features across domains is not considered. Simply put, all components except C, remain the same in the baseline model. The baseline, thus, evaluates the model performance in the absence of domain adaptation.

\renewcommand{\thefigure}{S\arabic{figure}}
\setcounter{figure}{0}

\begin{figure}[!ht]
    \centering
    \includegraphics[width=0.9\linewidth]{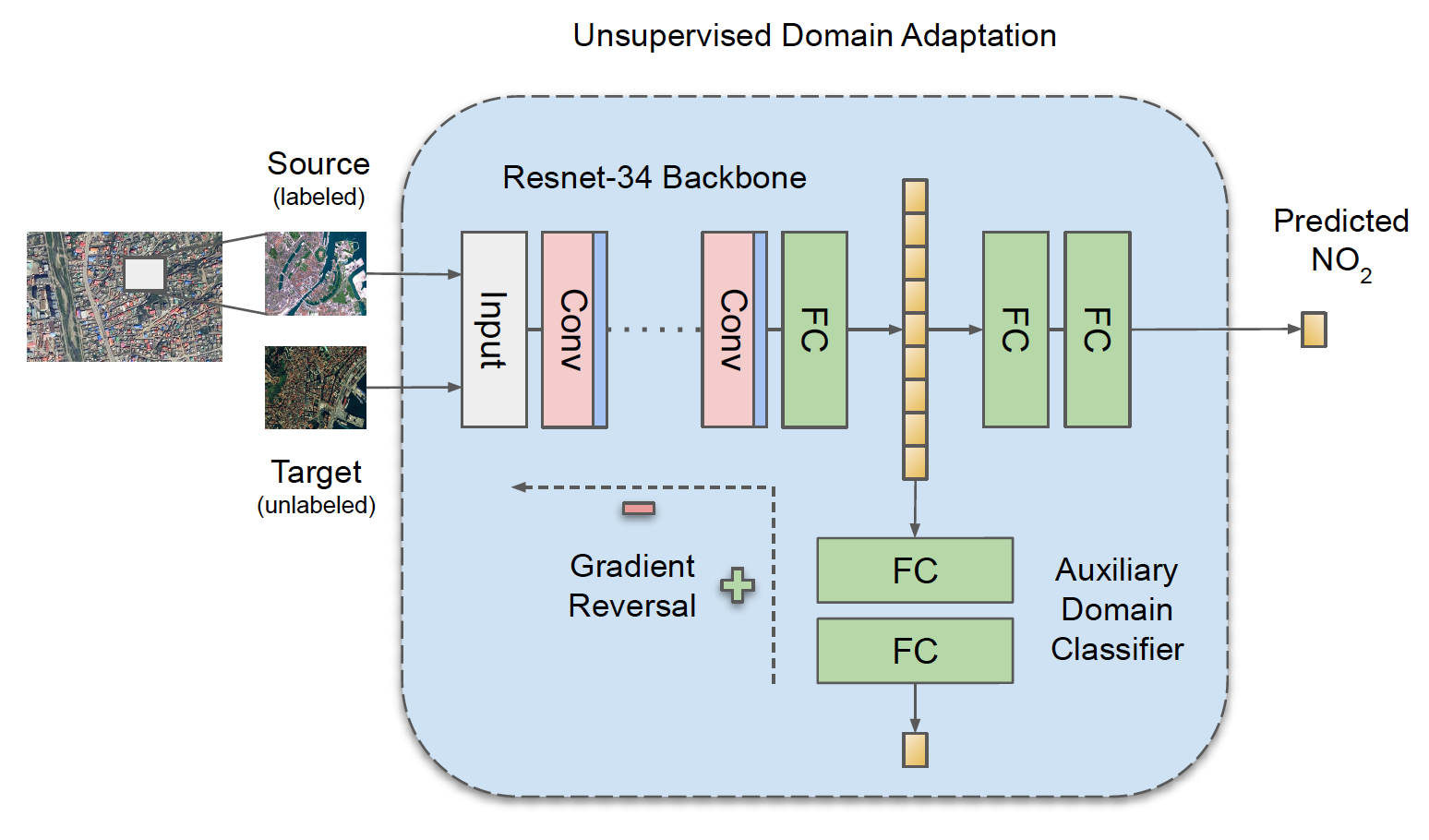}
    \caption{Schematic Architecture for DeepAQ - Unsupervised Domain Adaptation. The three main components include: a ResNet-34 feature encoder; an auxiliary domain classifier and a fully-connected decoder.}
    \label{fig:my_label}
\end{figure}

\begin{figure}[!ht]
\begin{center}
\includegraphics[width=1.0\linewidth]{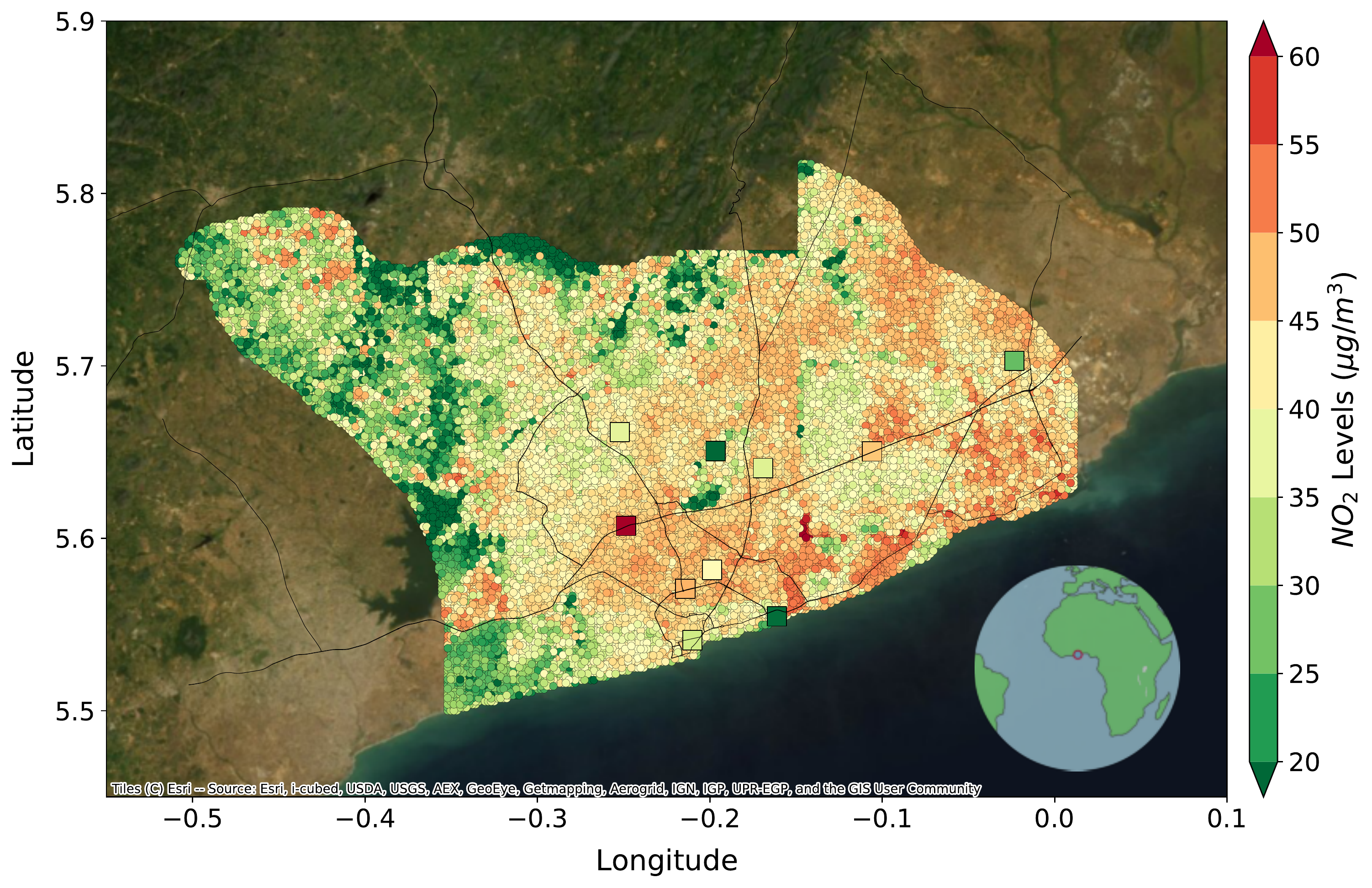}
\end{center}
   \caption{Mean annual $NO_2$ levels ($\mu g/m^3$) over Accra, Ghana, as predicted by DeepAQ. In this case, no road network information is provided to the model. The DeepAQ model is still able to capture the spatial distribution of $NO_2$ levels compared to the case where road network information is explicitly provided to the model.}
\end{figure}

\begin{figure}[!ht]
\begin{center}
\includegraphics[width=1.0\linewidth]{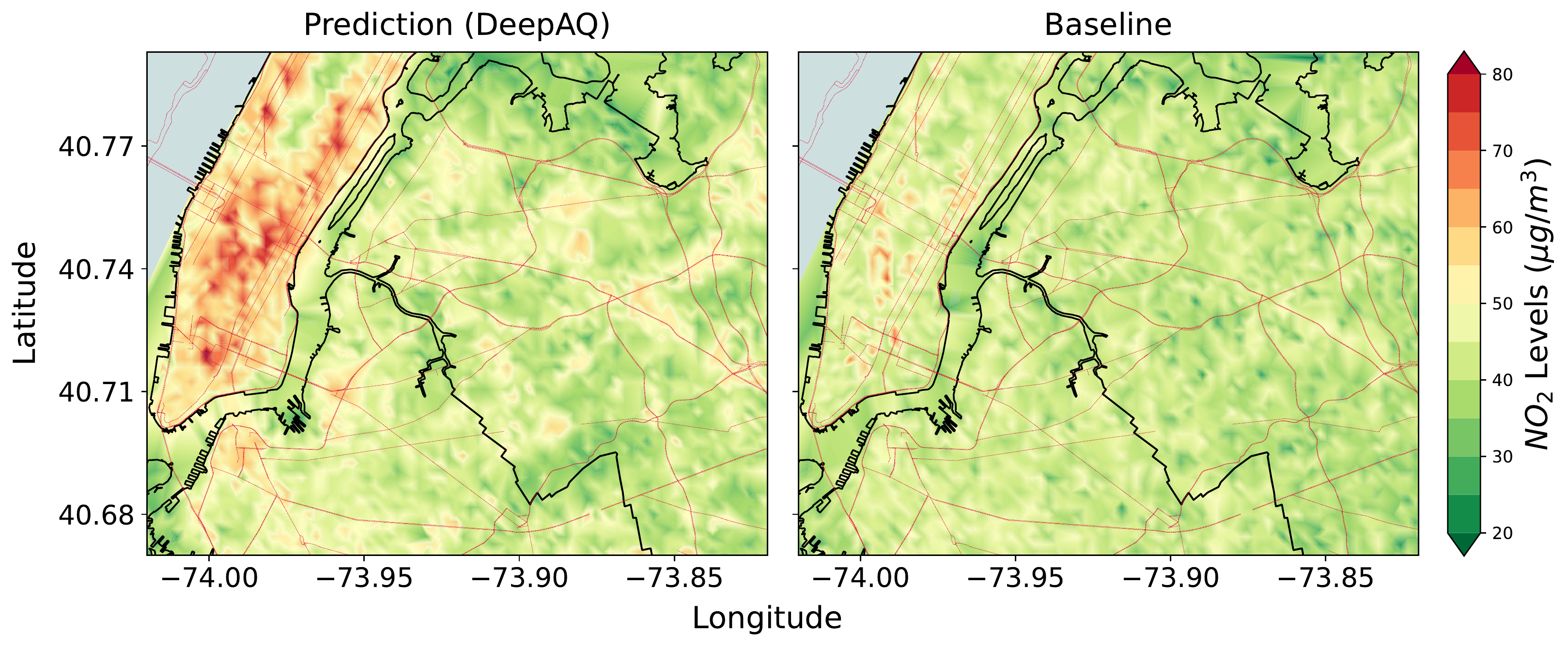}
\end{center}
   \caption{Mean annual $NO_2$ levels ($\mu g/m^3$) over New York City (NYC) as predicted by DeepAQ, and compared with the baseline Resnet-34 model (no domain adaptation). In this case, no road network information is provided to the model. The DeepAQ model is able to better identify regions of high $NO_2$ levels, such as in the Manhattan area of NYC (upper left); however, the performance is poorer compared to the case where road network information is provided explicitly. A possible explanation could be that $NO_2$ levels are highly correlated with proximity to roads in a large city like NYC where vehicular emission might be the predominant source of $NO_2$.}
\end{figure}

\begin{figure}[!ht]
\begin{center}
\includegraphics[width=0.7\linewidth]{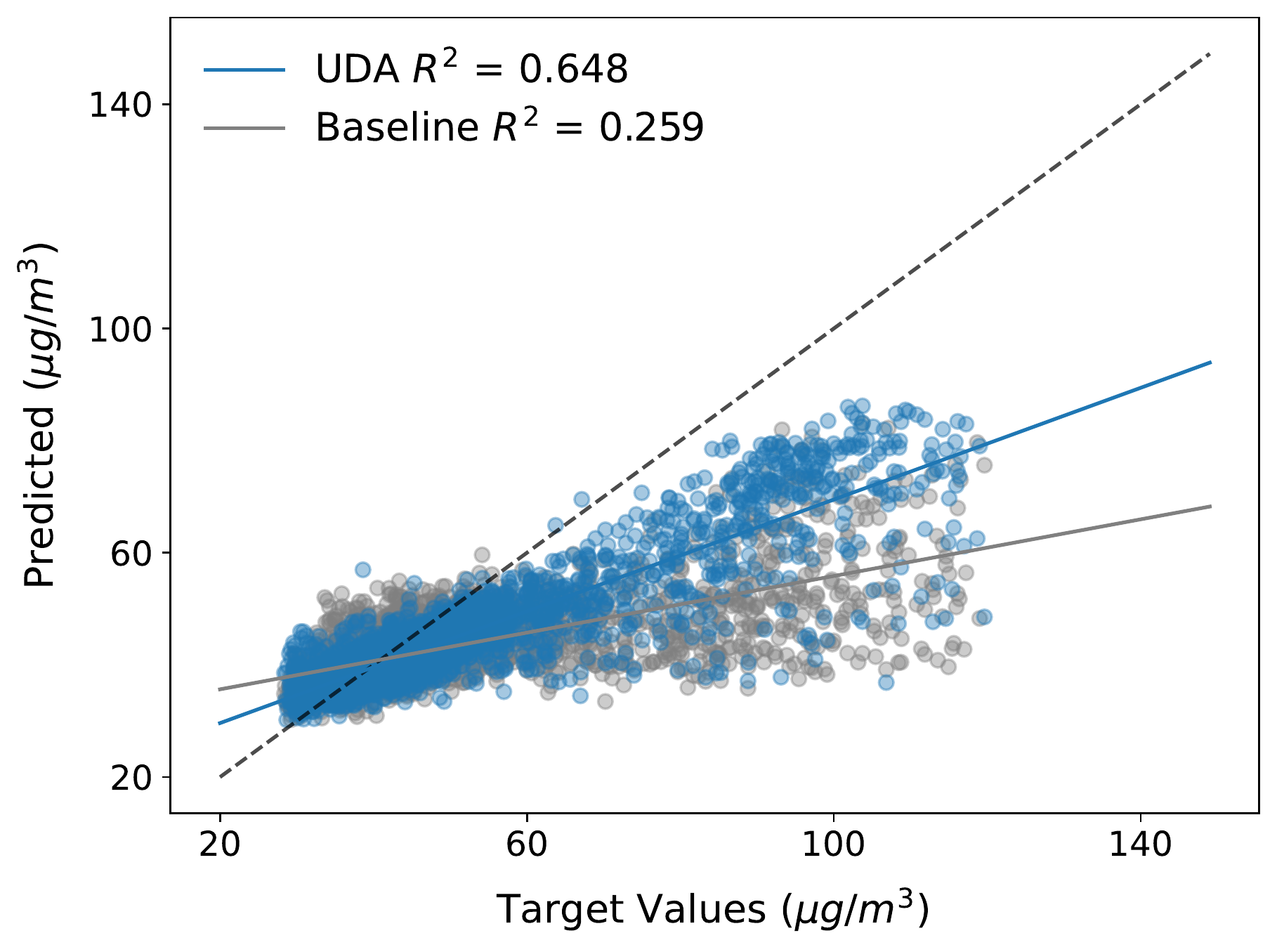}
\end{center}
   \caption{DeepAQ Predicted vs Target mean annual $NO_2$ ($\mu g/m^3$) distribution over New York City.}
\end{figure}

\begin{figure}[!ht]
\begin{center}
\includegraphics[width=0.7\linewidth]{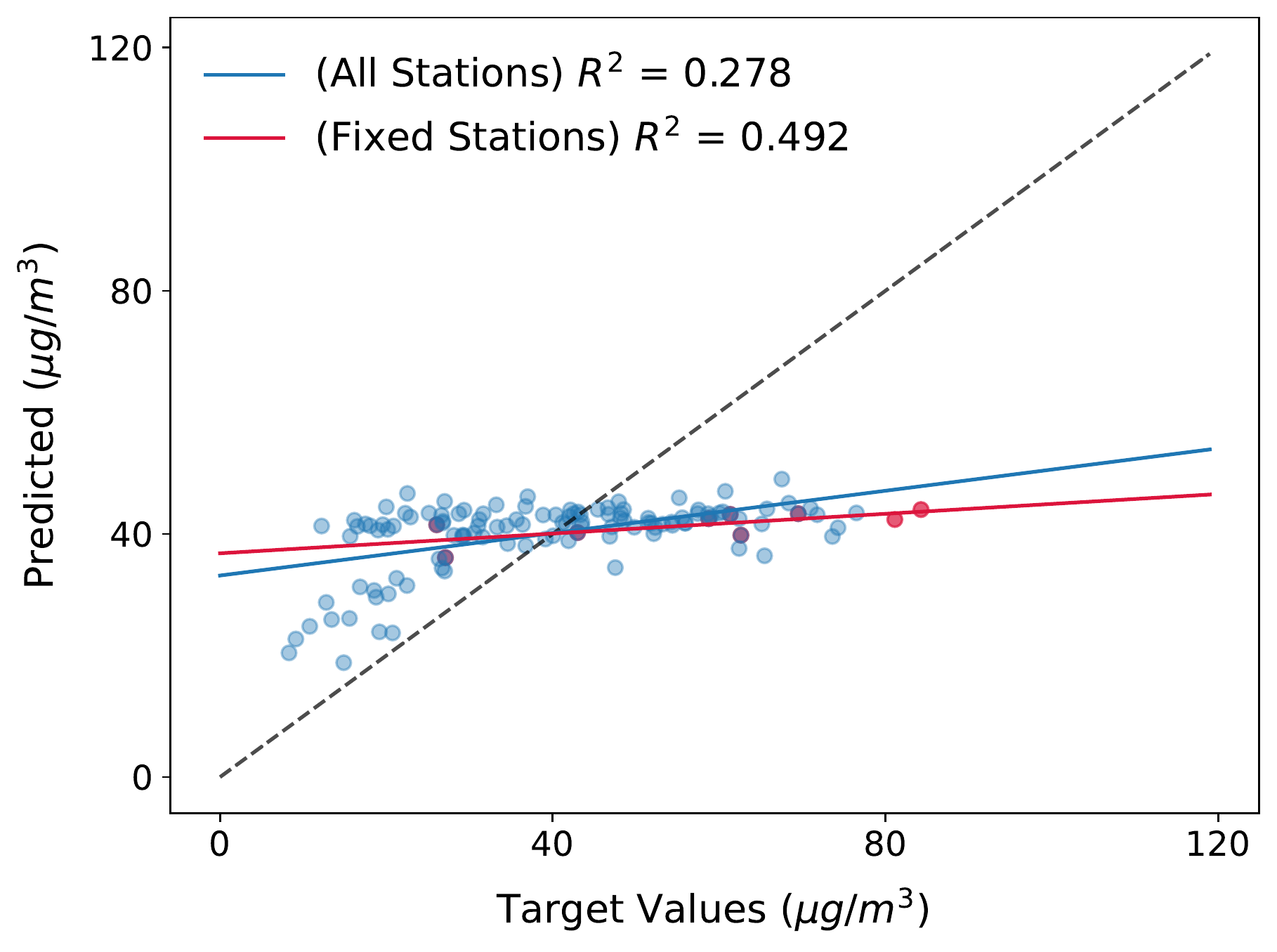}
\end{center}
   \caption{DeepAQ Predicted vs Target mean annual $NO_2$ ($\mu g/m^3$) distribution over Accra, Ghana.}
\end{figure}

\begin{figure}[!ht]
\begin{center}
\includegraphics[width=0.7\linewidth]{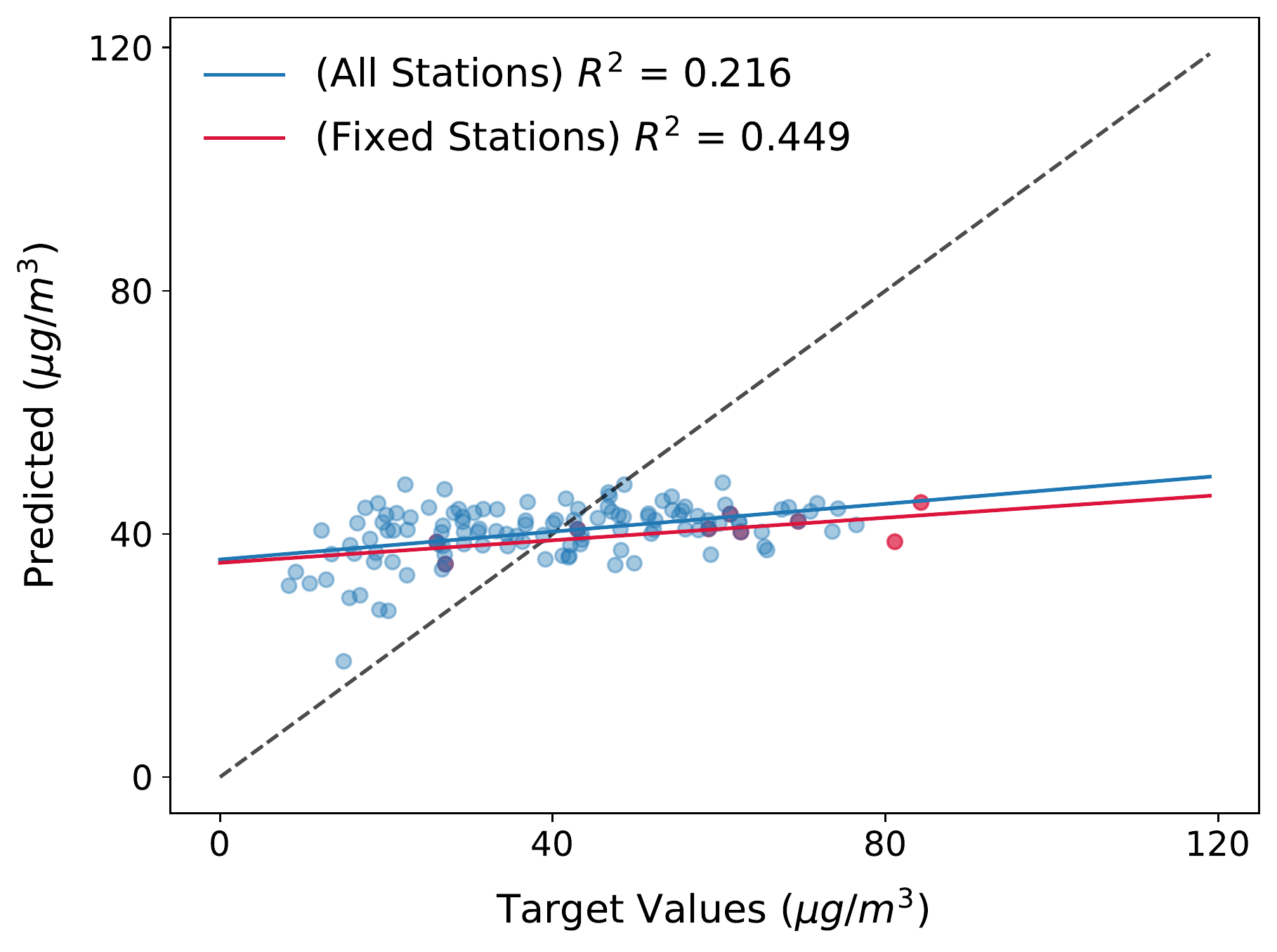}
\end{center}
   \caption{DeepAQ Predicted vs Targeted mean annual $NO_2$ ($\mu g/m^3$) distribution over Accra, Ghana when no road network information is provided to the DeepAQ model. For direct comparison, see figure S5.}
\end{figure}


\end{document}